\documentclass[letterpaper, 10 pt, conference]{ieeeconf}  

\usepackage{algorithmic}
\usepackage{epsfig}
\usepackage{graphicx}
\usepackage{amsmath}
\usepackage{amssymb}
\usepackage{booktabs}
\usepackage{multirow}
\usepackage{multicol}
\usepackage[ruled,vlined]{algorithm2e}
\usepackage{xcolor}
\usepackage{pifont}
\newcommand{\cmark}{\ding{51}}%
\newcommand{\xmark}{\ding{55}}%
\usepackage{xcolor}
\usepackage{csquotes}
\IEEEoverridecommandlockouts                              

\usepackage{pifont}
\usepackage{graphicx} 
\newcommand{\methodname}{ScaleDP}
\newcommand{\DP}{DP-T}
\usepackage{booktabs}                                   
\usepackage{multirow}                                   
\usepackage{makecell}                                   
\usepackage{tablefootnote}                              
\usepackage[symbol]{footmisc}                           
\usepackage{amsmath,amssymb}                            
\usepackage{xcolor}                                     
\let\labelindent\relax              
\usepackage{enumitem}                                   
\usepackage{subcaption}                                 
\usepackage{stfloats}                                   
\usepackage[misc]{ifsym}     
\usepackage{hyperref}   

\newcommand{\eat}[1]{}                                  

\title{\LARGE \bf
Scaling Diffusion Policy in Transformer to 1 Billion Parameters \\ for Robotic Manipulation
}

\author{Minjie Zhu$^{1,*}$, Yichen Zhu$^{2,*,\dagger}$, Jinming Li$^{5}$, Junjie Wen$^{1}$, Zhiyuan Xu$^{4}$,\\ Ning Liu$^2$, Ran Cheng$^2$,
Chaomin Shen$^{1,\dagger}$, Yaxin Peng$^{5}$, Feifei Feng$^{2}$, Jian Tang$^{4}$ 
\thanks{$^{1}$School of Computer Science, East China Normal University, China,
        {\tt\small \{mjzhu, jjwen, cmshen\}@cs.ecnu.edu.cn}}
\thanks{$^{2}$Midea Group, AI Research Center, 
        {\tt\small \{zhuyc25, liuning22, chengran, feifei.feng\}@midea.com}}
\thanks{$^{4}$Beijing Innovation Center of Humanoid Robotics, 
        {\tt\small \{eric.xu, jian.tang\}@x-humanoid.com}}
\thanks{$^{5}$Department of Computer Science, Shanghai University, China
        {\tt\small \{ljm2022, yaxin.peng\}@shu.edu.cn}}
\thanks{$*$ Co-first author. $\dagger$ Corresponding author.}
}

\begin{document}

\maketitle
\thispagestyle{empty}
\pagestyle{empty}

\begin{abstract}
Diffusion Policy is a powerful technique tool for learning end-to-end visuomotor robot control. It is expected that Diffusion Policy possesses scalability, a key attribute for deep neural networks, typically suggesting that increasing model size would lead to enhanced performance. However, our observations indicate that Diffusion Policy in transformer architecture (\DP) struggles to scale effectively; even minor additions of layers can deteriorate training outcomes. To address this issue, we introduce Scalable Diffusion Transformer Policy for visuomotor learning. Our proposed method, namely \textbf{\methodname}, introduces two modules that improve the training dynamic of Diffusion Policy and allow the network to better handle multimodal action distribution. First, we identify that \DP~suffers from large gradient issues, making the optimization of Diffusion Policy unstable. To resolve this issue, we factorize the feature embedding of observation into multiple affine layers, and integrate it into the transformer blocks. Additionally, our utilize non-causal attention which allows the policy network to \enquote{see} future actions during prediction, helping to reduce compounding errors. We demonstrate that our proposed method successfully scales the Diffusion Policy from 10 million to 1 billion parameters. This new model, named \methodname, can effectively scale up the model size with improved performance and generalization. We benchmark \methodname~across 50 different tasks from MetaWorld and find that our largest \methodname~outperforms \DP~with an average improvement of 21.6\%. Across 7 real-world robot tasks, our ScaleDP demonstrates an average improvement of 36.25\% over DP-T on four single-arm tasks and 75\% on three bimanual tasks. We believe our work paves the way for scaling up models for visuomotor learning. The project page is available at \href{scaling-diffusion-policy.github.io/}{https://scaling-diffusion-policy.github.io/}.
\end{abstract}
\section{Introduction}
Diffusion models have established leading roles in state-of-the-art advancements across various domains, including image, audio, video, and 3D generation~\cite{DDPM,DDIM,classifierddpm,Imagen,richter2023speech,sora,imagenvideo,lee2024dreamflow}. 
Specifically, Denoising Diffusion Probabilistic Models (DDPMs)~\cite{DDPM} are recognized for their approach of reversing a Stochastic Differential Equation. 
This technique leverages a stochastic denoising process that gradually incorporates Brownian motion during the generation of outputs.
Recently, the power of the diffusion model has manifested in the field of robotics as imitation learning~\cite{chi2023diffusion_policy,prasad2024consistency}.
It has become one of the most popular learning strategies for robotics, stimulating a series of improvements in skill learning, navigation, and visual representation.

The community expects that an effective method should be scalable: as the model size and training data increase, there should be a corresponding improvement in performance and generalization capabilities. 
This property, namely scaling laws, has driven remarkable progress across machine learning domains like language modeling~\cite{touvron2023llama, dehghani2023scaling} and computer vision~\cite{ha2023scaling, zhai2022scaling}, especially the success of large language models. 
Building a robot model that could achieve the scaling laws is also desirable in the field of robotics. However, whether Diffusion Policy (DP) could scale up, like those transformer models in other domains, has not been explored~\cite{tranformer,vit,kaplan2020scaling}. Hence, in this work, we study the scalability of the diffusion transformer for visuomotor policy learning.

\begin{figure*}[t]
  \centering
  \includegraphics[width=0.75\linewidth]{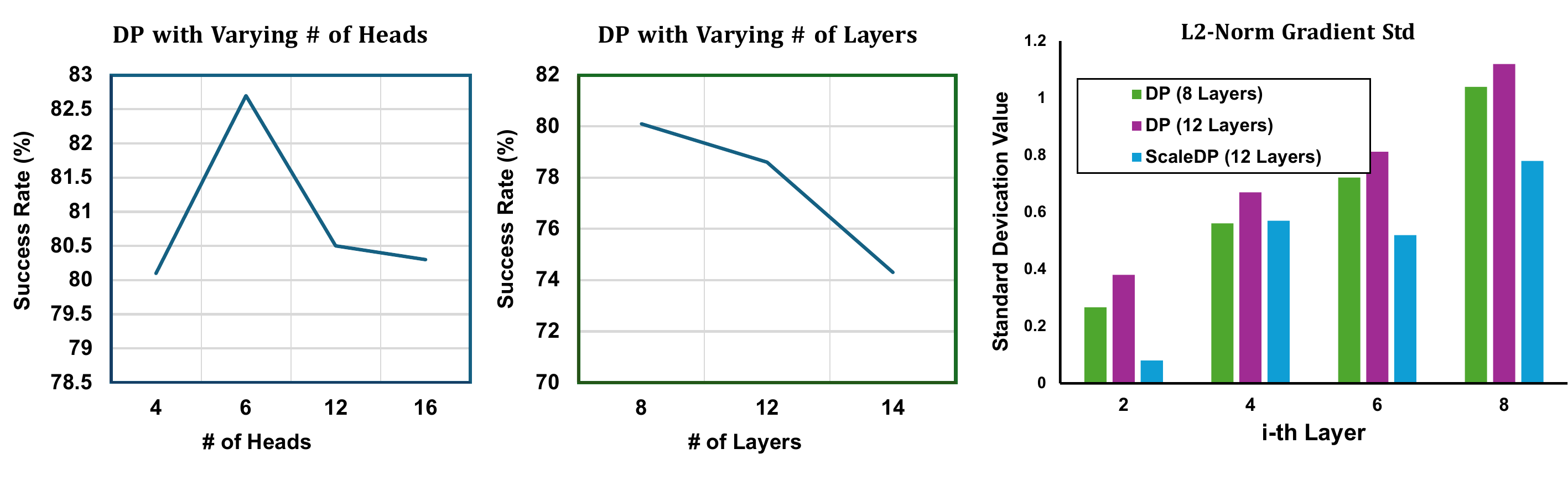}
  \caption{The motivation of \methodname. \textbf{Left:} Increasing the number of heads for Diffusion Policy in Transformer architecture does not necessarily improve performance. \textbf{Middle:} Increase depth could be harmful to the model performance. \textbf{Right:} The visualization of standard deviations of gradient magnitudes (the lower the better, i.e., more balanced optimization paces).}
  \label{fig:motivation}
\end{figure*}

We begin with the examination of the existing DP in transformer architecture (\DP). 
To assess the scalability of \DP, we conducted a preliminary investigation on MetaWorld~\cite{yu2020metaworld} (more details in Section~\ref{sec:motivation}). 
Our evaluation revealed that consistent with the findings in Diffusion Policy~\cite{chi2023diffusion_policy}, where scaling \DP~does not improve performance, regardless of increasing depth or number of heads, increasing in model size could negatively affect the tasks. 
For example (Figure~\ref{fig:motivation}), \DP~with eight layers achieves a success rate of 80.1\% in MetaWorld~\cite{yu2020metaworld}. 
However, this success rate decreases to 78.4\% when the number of layers is increased to twelve and further drops to 74.6\% with fourteen layers.

Through further analysis, we find that the failure to scale the transformer architecture stems from unstable training caused by large gradients in the observation fusion module. 
By replacing the conventional cross-attention fusion approach~\cite{tranformer} with multiple affine layers, we are able to normalize parameter distribution~\cite{chi2023diffusion_policy,karras2019style}, which brings good training dynamics to DP-T. 
To further improve model generalization, we propose to remove masked attention, allowing the model to \enquote{see} both past actions and future trajectories. 
This is particularly beneficial for learning visuomotor policies since trajectory predictions are typically much longer than the trajectories used during testing. 
For instance, Diffusion Policy predicts actions within ten timesteps but only uses the action at the first timestep. 
Allowing the model to observe the future trajectory makes it more robust to compound errors during prediction.

To demonstrate the effectiveness of our work, we conduct experiments on 50 simulation tasks in Metaworld and real robot experiments on 7 distinct tasks. 
We have successfully trained a Scalable Diffusion Transformer Policy (\textbf{\methodname}) that demonstrates effective scaling with an increase in model parameters, ranging from 10 million to 1 billion. 
Both simulation and real-world experiments reveal that \methodname~significantly outperforms the baseline Diffusion Policy. 
Additionally, we confirm that as the model size increases, it accommodates more training data, which enhances its performance. 
Furthermore, we observe improved visual generalization capabilities, as the model scales up.

%


\section{Related Works}

\noindent
\textbf{Diffusion Policy in Robotics.} Diffusion models are one of the generative models that progressively transform random noise into structured data samples, have achieved remarkable success in generating high-fidelity images~\cite{DDPM,DDIM,stablediffusion,Imagen}. Due to their remarkable expressiveness, diffusion models have recently expanded into the field of robotics. Their applications now extend to various domains such as reinforcement learning~\cite{yang2023policy,mazoure2023value,brehmer2024edgi,venkatraman2023reasoning,lee2024refining,zhou2024adaptive,chen2022offline,ze2023visual}, imitation learning~\cite{chi2023diffusion_policy,prasad2024consistency,khazatsky2024droid,vosylius2024render,reuss2024multimodal,ze20243d,team2024octo,xian2023unifying,ze2023gnfactor,ha2023scaling,reuss2023goal,pearce2023imitating,ze2023visual3d,zhu2024retrieval,ze2023gnfactor,yang2024movie}, reward learning~\cite{psenka2023learning,huang2023diffusion,ze2024h}, grasping~\cite{urain2023se, simeonov2023shelving, wu2024learning}, and motion planning~\cite{wen2024tinyvla,zhong2023language,urain2023se,liu2023mm,zhu2024llava,liu2022structdiffusion,saha2023edmp,chang2023denoising,wen2024object,zhu2024language}.
In this work, we focus on scaling up the diffusion policy with transformer architectures.
We demonstrate that Diffusion Policy in transformer architecture fails to scale up. We show that our proposed method, when scaled up, gain possesses multiple merits that the small transformer model does not have.

\section{Method}
\noindent
\textbf{Problem Setup.} We assume an expert collected dataset of demonstrations $\mathcal{D} = \{\tau_0, \tau_1, \ldots, \tau_n\}$, where each trajectory $\tau_i = \{(o_j, x_j)\}$ is a sequence of paired raw visual observations $o$ and proprioceptive information $x$. The proprioceptive information can either be the end-effector pose or joint angles and includes the gripper width. In this work, we use 6D pose, i.e., position $(x, y, z)$ and rotation $(roll, pitch, yaw)$ to control the robot.

\noindent
\textbf{Diffusion Policy.} Diffusion Policy \cite{chi2023diffusion_policy} models the conditional action distribution as a denoising diffusion probabilistic model (DDPM) \cite{DDPM}, allowing for better representation of the multi-modality in human-collected demonstrations. 
Specifically, Diffusion Policy uses DDPM to model the action sequence $p(A_t \mid o_t, x_t)$, where $A_t = \{a_t, \ldots, a_{t+C}\}$ represents a chunk of the next $C$ actions. 
The final action is the output of the following denoising process~\cite{welling2011bayesian}:

\begin{equation}
A_t^{k-1} = \alpha \left( A_t^k - \gamma \epsilon_\theta (o_t, x_t, A_t^k, k) \right) + \mathcal{N}(0, \sigma^2 I),
\label{eqn1}
\end{equation}

where $A_t^k$ is the denoised action sequence at time $k$. 
Denoising starts from $A_t^K$ sampled from Gaussian noise and is repeated until $k = 1$. 
In Eqn (\ref{eqn1}), $(\alpha, \gamma, \sigma)$ are the parameters of the denoising process and $\epsilon_\theta$ is the score function trained using the MSE loss $\ell(\theta) = (\epsilon_k - \epsilon_\theta (o_t, x_t, A_t^k + \epsilon_k, k))^2$. 
The noise at step $k$ of the diffusion process, $\epsilon_k$, is sampled from a Gaussian of appropriate variance.

\subsection{Example of Motivation}
\label{sec:motivation}
To better illustrate the scalability problem of Diffusion Policy, we leverage MetaWorld~\cite{yu2020metaworld} as our testbed. The experimental results are presented in Figure~\ref{fig:motivation}. Our findings indicate that increasing the model size of the vanilla Diffusion Policy in Transformer architecture (\DP)~\cite{chi2023diffusion_policy} does not consistently enhance the success rate on tasks in MetaWorld. This observation is consistent with the statement made in the original Diffusion Policy paper~\cite{chi2023diffusion_policy}. Specifically, there is a noticeable performance boost when the number of heads increases from four to six. However, adding more heads beyond this point results in the average success rate reverting to that of a model with only four heads.

We also assessed the impact of increasing the number of layers within the Transformer model. Our empirical results show a consistent decline in performance with each additional layer. For example, a model with eight layers achieves a success rate above 80\%, but this decreases to 78.4\% with twelve layers and drops below 75\% with fourteen layers. 

These findings suggest that the current Diffusion Policy model struggles to scale effectively with respect to model size. This scalability limitation hampers the model's ability to learn from data, ultimately diminishing its generalization capabilities. We further investigated the training dynamics of \DP, plotting the standard deviation of gradient magnitudes across different layers. Previous works~\cite{zhao2022penalizing, zhou2020towards, pmlr-v139-akbari21a} found that lower values indicate a more balanced optimization pace, which generally leads to better generalization. As illustrated in Figure~\ref{fig:motivation} (right), increasing the depth of \DP~results in larger gradient magnitudes, signaling unstable training in deeper network configurations. This motivated us to modify the neural architecture to address this issue. We demonstrate that \methodname~maintains low gradient magnitudes even with an increased number of layers.

\begin{figure}[tb]
  \centering
  \includegraphics[width=1\linewidth]{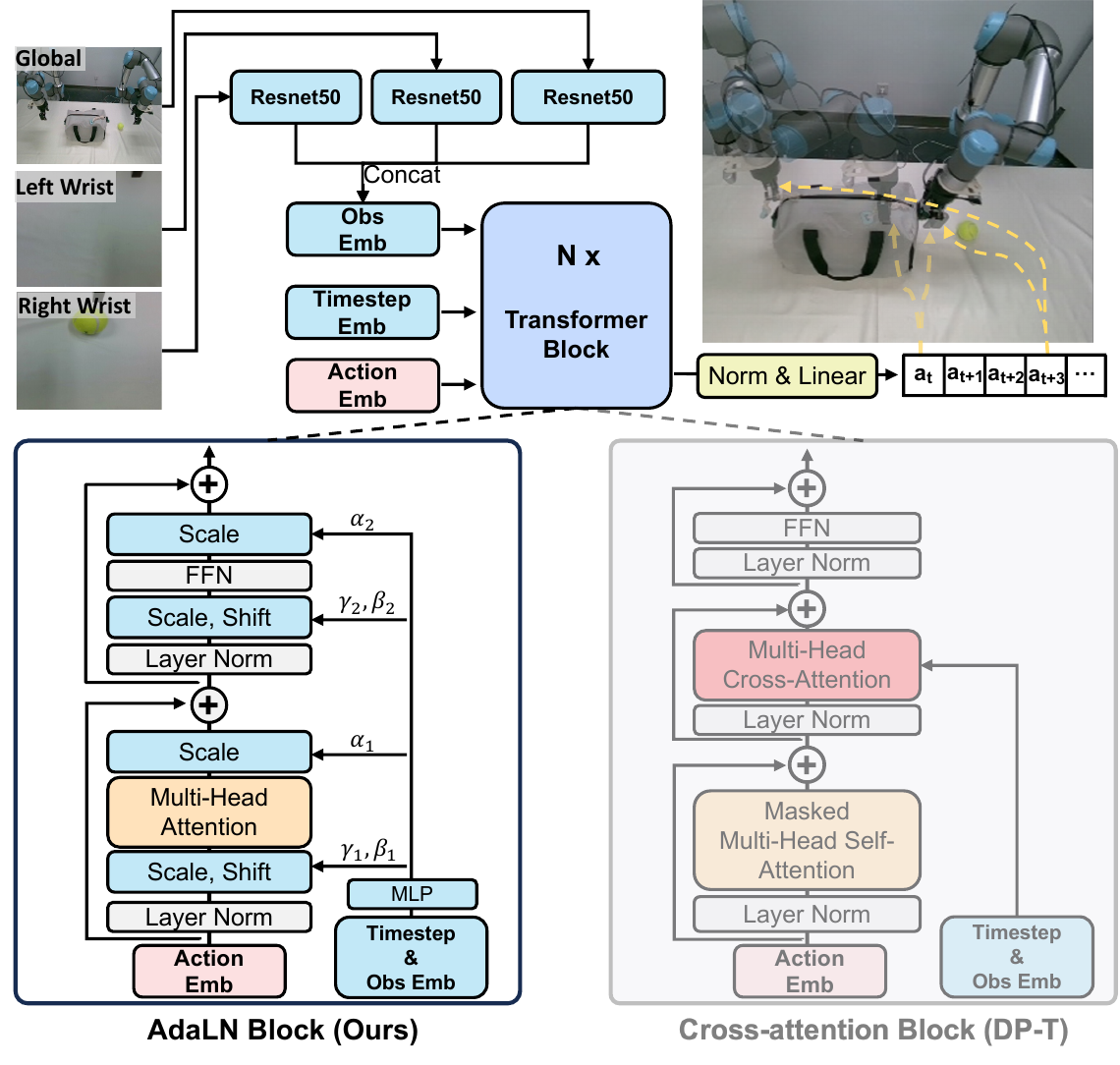}
  \caption{\textbf{The architecture of the scalable diffusion transformer policy.} \textit{Top:} Overview of Our \methodname. It takes as input multi-view images and outputs a sequence of actions. \textit{Bottom:} Details of our \methodname~block. The cross-attention block has the same structure as \cite{tranformer}. The \textbf{AdaLN block} employs adaptive layer norm to fuse conditions into the noise action embeddings, achieving more stable training and better inference performance.}
  \label{fig:example}
\end{figure}
\subsection{Modification on Neural Architecture}
This section gives a detailed illustration of how we modify our neural architecture to ensure \methodname~could scale up the model size.

\noindent
\textbf{Cross-attention block.} The traditional approach fuses the conditional information with a cross-attention mechanism~\cite{chi2023diffusion_policy,tranformer}.
It concatenates the embeddings of timestep $k$ and observation $o$ into a sequence, separate from the action sequence. 
The transformer block is similar to the original design from~\cite{tranformer}. 
We find that increasing the depth of \DP~results in larger gradient magnitudes, thus making the training procedure more difficult.

\noindent
\textbf{Adaptive Layer Norm (AdaLN) block.} Following the widespread usage of adaptive normalization layers in image generation~\cite{karras2019style,peebles2023scalable,brock2018large}, we explore replacing standard layer norm layers with adaptive layer norm (AdaLN).
Specifically, instead of directly learning dimension-wise scale and shift parameters $\gamma$ and $\beta$, we regress them from the sum of the embedding vectors of $k$ and $o$. 
Compared with the conditioning mechanism using cross-attention, this enables the model to change the distribution of the noise action embedding according to the conditions. 
The AdaLN is defined as:
\begin{equation}
\text{AdaLN}(x) = (\gamma(k, o) + 1) \cdot x + \beta(k, o)
\end{equation}
where $x$ is the input to the layer normalization, and $\gamma(k, o)$ and $\beta(k, o)$ are the adaptive scale and shift parameters regressed from the embedding vectors of $k$ and $o$.

\noindent
\textbf{Non-causal Attention.}  Following the transformer architecture proposed by \cite{tranformer}, the Diffusion Policy utilizes masks to ensure that each action embedding can only attend to previous tokens in the self-attention and cross-attention layers of each transformer decoder block.
We argue that this unidirectional attention mechanism would hide the action representations. By removing the mask in self-attention layers, we can make each action more consistent with both left and right actions.
 
We apply a sequence of $N$ \methodname~blocks, each operating at the hidden dimension size $d$. \
Following ViT, we use standard transformer configs that jointly scale $N$, $d$,  and attention heads~\cite{vit}. Specifically, we use five configs: \methodname-Ti, \methodname-S, \methodname-B, \methodname-L, and \methodname-H. 
They cover a wide range of model sizes, from 10M parameters to 1B parameters, allowing us to gauge scaling performance. Table~\ref{tab:config} gives details of the configs.

After the final \methodname~block,
we apply the final adaptive layer norm and linear layer to decode the sequence of noise action tokens into the predicted noise.

\section{Experiments}
\begin{figure}[tb]
  \centering
  \includegraphics[width=\linewidth]{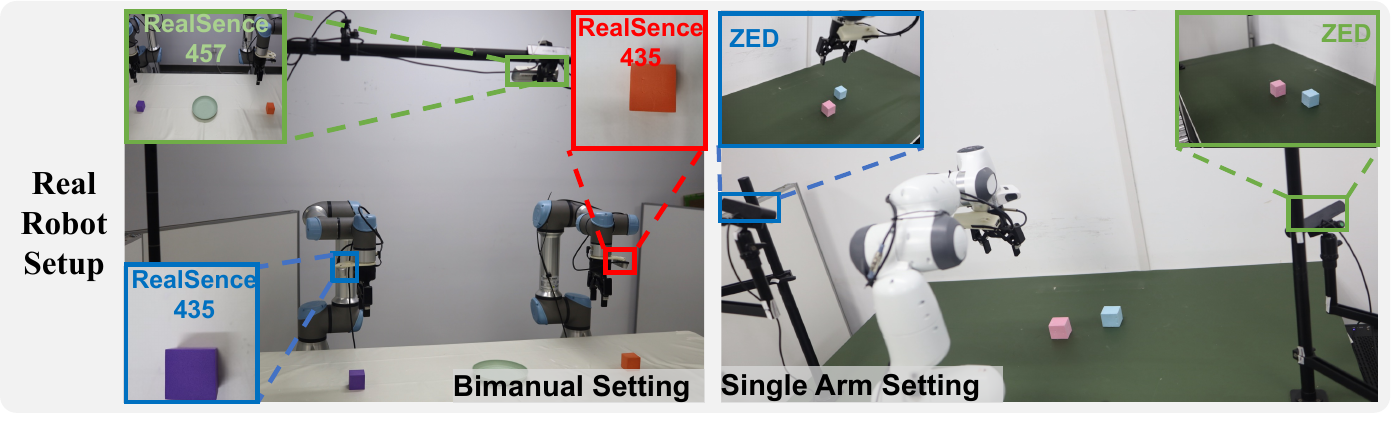}
  \caption{To evaluate the wide scalability of \methodname, we conduct experiments on both the Bimanual UR5 Robot Arms and the Franka Arm.}
  \label{fig:workspace}
\end{figure}

\begin{table}[t]
\centering
\caption{Diverse model size of \methodname. We present five model sizes, Tiny (Ti), Small (S), Base (B), Large (L), and Huge (H).}
\label{tab:config}
\resizebox{0.45\textwidth}{!}{\begin{tabular}{c|c|c|c|c}
\toprule
\textbf{Model: } & \multirow{2}{*}{\textbf{Layers}} & \textbf{Hidden size} & \multirow{2}{*}{\textbf{Heads}} & \textbf{Param} \\
\methodname & & $d$ & & \\
\midrule
Tiny (Ti) & 8 & 256 & 4 & 10M \\ 
Small (S) & 12 & 384 & 6 & 33M \\ 
Base (B) & 12 & 768 & 12 & 130M\\ 
Large (L) & 24 & 1024 & 16 & 457M \\ 
Huge (H) & 32 & 1280 & 16 & 1B \\ 
\bottomrule
\end{tabular}}
\end{table}

In our experiments, we aim to demonstrate the effectiveness of ScaleDP from the following two perspectives: 1) The performance compared to Diffusion Policy, the model/data scalability, and rate of convergence; 2) The visual observation appears on the model, including appearance, object, light, and distractor. 
\begin{figure*}[t]
  \centering
  \includegraphics[width=0.9\linewidth]{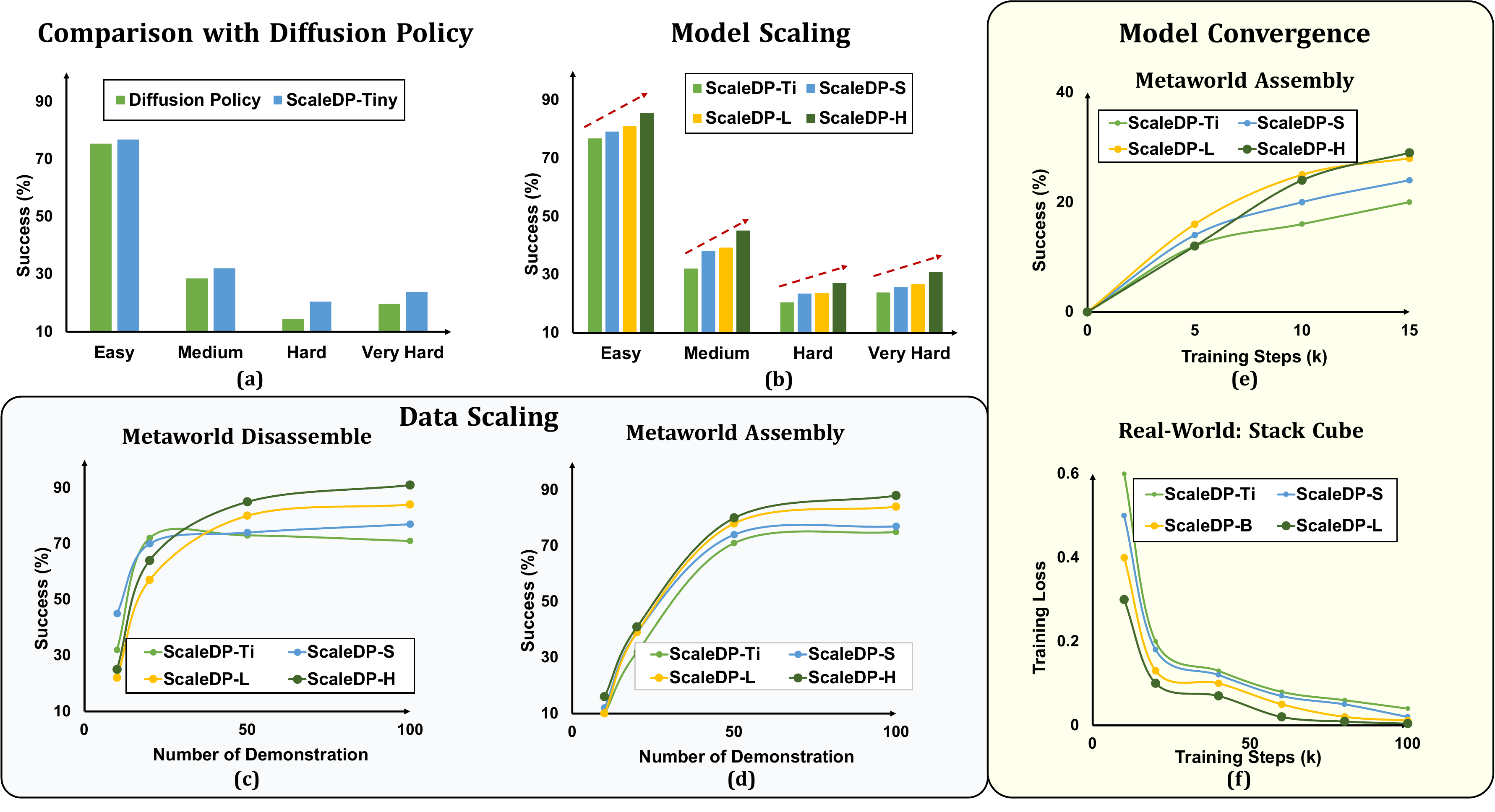}
  \caption{\textbf{Experiments.} (a) Comparison with Diffusion Policy on MetaWorld; (b) Model scaling results on MetaWorld; (c) and (d): Data scaling results on MetaWorld (Disassemble and Assembly); (e) Model convergence rate on MetaWorld Assembly; (f) Model convergence rate on real-world task (stack cube).}
  \label{fig:simulation}
\end{figure*}

\subsection{Real Robot Experimental Setup}

\noindent
\textbf{Real robot benchmarks.} Our \methodname~is evaluated across 7 tasks, with 4 tasks using Franka robot with a 7-degree-of-freedom arm and 3 tasks using two UR5 robots with a total of 14-degree-of-freedom arm. We use 2 ZED cameras for Franka and 3 RealSence cameras for bimanual to obtain real-world visual information. 
Our real robot setup are shown in Figure~\ref{fig:workspace}. A brief description of our tasks is following:

\noindent
\textbf{Data collection.} We acquire our dataset through demonstrations performed by humans. For each target task, we place objects randomly within a designated area and instruct a human to manipulate the objects as smoothly as possible. Additionally, the opening of the mug faces to the left for flip mug task. Throughout this process, we record the RGB streams from two different angles and capture the robot's state, such as joint positions. \methodname~employs a mainstream control mode that predicts the 6D, encompassing position $(x, y, z)$ and rotation $(roll, pitch, yaw)$. For every task, we collected 100 trajectories. For the task of closing a laptop, we collected 40 trajectories.

\noindent
\textbf{Baselines.} As the primary focus of this work is to study the scalable diffusion transformer policy, we select the vanilla Diffusion Policy in transformer architecture (\DP)~\cite{chi2023diffusion_policy} as our baseline. We also compare a number of different approaches, including Octo~\cite{team2024octo}, Beso~\cite{reuss2023goal}, MDT~\cite{reuss2024multimodal}, DP-Unet~\cite{chi2023diffusion_policy} and ACT~\cite{act}. These methods cover models with transformer architecture and other variants of Diffusion Policy. 

\subsection{Simulation Experiments}

\noindent
\textbf{Experimental setup.} We classified 50 tasks from MetaWorld~\cite{yu2020metaworld} into levels—easy, medium, hard, and very hard—based on MWM~\cite{seo2023masked}. All experiments were trained with 20 demonstrations and evaluated with 3 seeds, and for each seed, the success rate was averaged over five different iterations.

\noindent
\textbf{Comparison with \DP.} We compared the \methodname-Ti model with \DP. Both models have a comparable number of parameters. As shown in Figure~\ref{fig:simulation}(a), our approach achieves a higher success rate across all four levels of challenging tasks in MetaWorld. Notice that \methodname-Ti has a similar number of parameters as the Diffusion Policy. The superior performance of \methodname-Ti across all task levels indicates a more efficient utilization of the model's capabilities, due to more advanced architecture as we proposed.

\noindent
\textbf{Model scaling.} In Figure~\ref{fig:simulation}(b), we present the results of scaling up the model size while keeping the number of demonstrations constant. The data indicate that as the model size increases, the success rate improves, demonstrating the scalability of our method. This pattern confirms that our approach not only accommodates but thrives of increased computational capacity. The continuous improvement in success rates with larger model sizes, despite the constant number of demonstrations, suggests that the models are effectively extracting more meaningful patterns and insights from the same amount of data.
\\
\\
\noindent
\textbf{Data scaling.} We further explored whether larger models benefit more from increased data. Figures~\ref{fig:simulation} (c) and (d) show that as the number of demonstrations increases, the success rate for smaller models plateaus, whereas larger models continue to improve. This trend suggests that larger ScaleDP have a higher capacity to leverage additional data, thereby enhancing their learning curves significantly. Moreover, this observation underscores the importance of data scalability when deploying larger models in practical applications.

\noindent
\textbf{Learning efficacy.} To illustrate the learning efficacy of our model, we plotted the model convergence in Figures~\ref{fig:simulation} (e) and (f). Figure~\ref{fig:simulation} (e) shows the success rate on the MetaWorld Assembly task, while Figure~\ref{fig:simulation} (f) examines the training loss on a real robot task. The results indicate that as training progresses, larger models tend to converge more effectively, achieving higher success rates and lower training losses.


\begin{table}[t]
\centering
\caption{Success rates on \textbf{four real-world tasks on single arm Franka robot}. Task 1: Close Laptop, Task 2: Flip Mug, Task 3: Stack Cube, Task 4: Place Tennis. It is worth noting that as the model size increases, the average success rate also increases correspondingly, demonstrating the scalability of our model architecture. Each task is evaluated with 20 trials.}
\label{tab:real-exp-single}
\resizebox{0.45\textwidth}{!}{\begin{tabular}{c|c|c|c|c|c}
\toprule
\textbf{Model}                      & Task1 & Task2 & Task3 & Task4 & \textbf{Avg.} \\
\midrule
Octo~\cite{team2024octo} & 65 & 50 & 40 & 35 & 47.50\\
Beso~\cite{reuss2023goal} & 50 & 30 & 20 & 15 & 28.75\\
MDT~\cite{reuss2024multimodal} & 55 & 45 & 50 & 30 & 45.00\\
DP-Unet~\cite{chi2023diffusion_policy} & 70 & 70 & 45 & 40 & 56.25\\
ACT~\cite{act}                      & 90 & 70 & 55 & 50 & 66.25 \\
DP-T~\cite{chi2023diffusion_policy} & 80 & 70 & 50 & 5  & 51.25 \\  
\midrule
\methodname-S                       & 85 & 70 & 50 & 30 & 58.75 \\ 
\methodname-B                       & 80 & 65 & 50 & 55 & 62.50 \\ 
\methodname-L                       & \textbf{95} & 80 & 70 & 50 & 73.75 \\ 
\methodname-H                       & \textbf{95} & \textbf{95} & \textbf{90} & \textbf{70} & \textbf{87.50} \\ 
\bottomrule
\end{tabular}}
\end{table}

\begin{table}[t]
\centering
\caption{Success rates on \textbf{three real-world tasks on bimanual UR5 robot}. Task 1: Put tennis ball into bag, Task 2: Sweep trash, Task 3: Bimanual Stack Cube. It is worth noting that as the model size increases, the average success rate also increases correspondingly, demonstrating the scalability of our model architecture. Each task is evaluated with 20 trials.}
\label{tab:real-exp-biamnual}
\resizebox{0.4\textwidth}{!}{\begin{tabular}{c|c|c|c|c}
\toprule
\textbf{Model} & Task1 & Task2 & Task3 & \textbf{Avg.} \\
\midrule

ACT~\cite{act}                                          & \textbf{100} & 70	& 50 & 73.33  \\
DP-T~\cite{chi2023diffusion_policy}                    & 20 & 50 & 0 & 23.33\\ 
\midrule
\methodname-S &  \textbf{100} & 50 & 10 & 53.33  \\ 
\methodname-B &        \textbf{100} & 60 & 10 & 56.67  \\ 
\methodname-L    & \textbf{100} & 80 & 90 & 90.00 \\ 
\methodname-H    & \textbf{100} & \textbf{95} & \textbf{100} & \textbf{98.33} \\ 
\bottomrule
\end{tabular}}
\end{table}

\begin{table}[t]
\centering
\caption{Ablation study on the effectiveness of non-causal attention on real-world tasks. The experiments are conducted on bimanual UR5 robot.}
\label{tab:unmask}
\resizebox{0.47\textwidth}{!}{\begin{tabular}{c|c|c|c|c|c}
\toprule
\textbf{Model} & Non-causal & Task1 & Task2 & Task3 & \textbf{Avg.} \\
\midrule
\multirow{2}{*}{\methodname-S} & \xmark                      & 70 & 20 & \textbf{10} &  33.33                                     \\
                               & \cmark                     & \textbf{100} & \textbf{50} & \textbf{10} & \textbf{$53.33_{\textcolor{blue}{+20}}$}  \\ 
                               \midrule
\multirow{2}{*}{\methodname-B} & \xmark                       & 90 & 50 & \textbf{10} & 50.00 \\
                               & \cmark                     & \textbf{100} & \textbf{60} & \textbf{10} & \textbf{$56.67_{\textcolor{blue}{+6.67}}$}  \\ 
                               \midrule
\multirow{2}{*}{\methodname-L} & \xmark                       & \textbf{100} & \textbf{80} & 20 & 66.66 \\
                               & \cmark                     & \textbf{100} & \textbf{80} & \textbf{90} & \textbf{$90.00_{\textcolor{blue}{+23.34}}$} \\ 
\bottomrule
\end{tabular}}
\end{table}

\subsection{Real Robot Experiments}

\noindent
\textbf{Main result.} Table~\ref{tab:real-exp-single} and ~\ref{tab:real-exp-biamnual} presents the real robot experimental results. It can be observed that \methodname~outperforms \DP~across all model sizes in 3 bimanual tasks and 4 single-arm tasks. Notably, for place tennis task, \DP~succeeded only once in 20 trials, whereas our \methodname-B/L achieved a success rate of at least 50\%. Moreover, as the model size increases, the average success rate improves correspondingly, demonstrating the scalability of our model architecture. Additionally, compared with state-of-the-art imitation learning method, such as ACT, \methodname-L outperform its average success rates by 16.67\% on 3 bimanul tasks and by 7.5\% on 4 single-arm tasks, while \methodname-S and \methodname-B do not, further highlighting the scalability of our \methodname's architecture. When we increase the model size to 1 billion parameters, ScaleDP-H achieves even better performance across all tasks and experimental settings. It improves the average success rate over ScaleDP-L by 13.75\% and 8.33\% in two different setups, respectively. These results validate the scalability of our method and highlight the importance of increasing model size in diffusion-based imitation learning.

\noindent
\textbf{Non-causal attention.} To demonstrate the importance of the non-causal attention in \methodname, we conducted ablation studies on \methodname~across 3 bimanual tasks (see Table~\ref{tab:unmask}). Our findings indicate that unmasking significantly improves the test performance of all 3 model sizes, particularly for the large model size, which shows a remarkable improvement on Task 3 (Bimanual Stack Cube), achieving a success rate that is 70\% higher than that with masking strategy.

\subsection{Visual Generalization}
\begin{figure*}[t]
  \centering
  \includegraphics[width=0.9\linewidth]{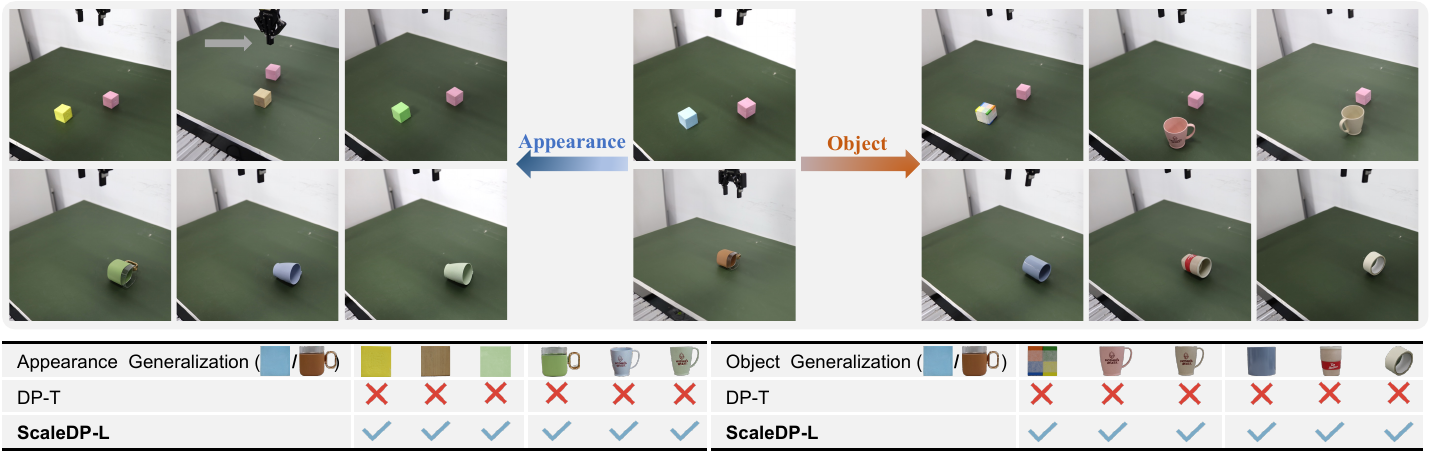}
  \caption{\textbf{Appearance \& Object Generalization.} We test two tasks: stack cube and flip mug. For appearance generalization, we change the color of the target object to another color. For object generalization, we replace the target object with six objects of varied sizes and shapes from daily life. Each result is evaluated with one trial.}
  \label{fig:aog}
\end{figure*}

\begin{figure*}[tb]
  \centering
  \includegraphics[width=0.9\linewidth]{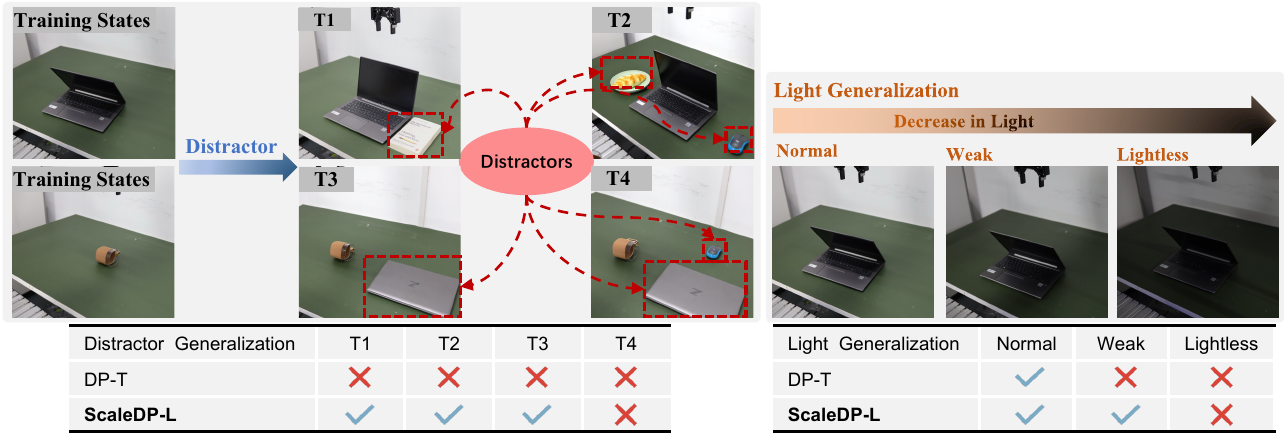}
  \caption{\textbf{Distractor \& Light Generalization.} We test these capabilities on close laptop and flip mug tasks. Each test is evaluated with one trial.}
  \label{fig:dlg}
\end{figure*}

Visual generalization refers to the ability to adapt to novel visual textures. Examples of this in robotic manipulation tasks include variations in background color, object texture, or ambient lighting. These visual changes do not alter the fundamental structure of the task, such as the positions of objects and targets, and primarily require the robot to accurately interpret semantic meanings. Here we demonstrate the visual generalization ability of \methodname-L. We categorize the visual generalization into the following:

\noindent
\textbf{Appearance generalization.} 
We alter the color of the target objects to be grabbed, as demonstrated in Figure~\ref{fig:aog}. Originally, the cube/mug is colored blue/gold; we then make changes accordingly. We observe that \methodname-L~is able to generalize on objects with different colors. In comparison, the vanilla DP fails to recognize target objects of different colors. Notably, our approach achieves appearance generalization without relying on data augmentation during training. This indicates that the generalization capability of our model stems solely from its ability to recognize the shapes of objects.

\noindent
\textbf{Object generalization.} Achieving generalization across diverse objects, which vary in size and shape, presents a significantly greater challenge compared to mere appearance generalization. In Figure~\ref{fig:aog}, we demonstrate that \methodname-L~effectively manages a wide range of everyday objects. Specifically, when the blue block is replaced with a cup of a completely different shape and the gold mug with a tape, \methodname-L~still exhibits robust generalization capabilities. This ability to adapt to new and varied object types without a loss in performance underscores the flexibility and practicality of our model, making it highly suitable for dynamic and unpredictable real-world environments where object variability is the norm.

\noindent
\textbf{Light generalization.} Light generalization is similar to background generalization, which reduces the intensity of the light background in each image compared to the normal one. As demonstrated in Figure~\ref{fig:dlg}, \methodname-L~effectively addresses this generalization problem when the light is slightly decreased. However, it is crucial to acknowledge that while \methodname-L~can generalize across minor variations in light, significant changes might be more difficult to handle.

\noindent
\textbf{Distractor generalization.} Distractor generalization refers to introducing additional distractors during the testing phase to evaluate a model's ability to resist distractions. As shown in Figure~\ref{fig:dlg} (T4), simply adding a mouse to the scene, compared with T3, prevents \methodname-L~from completing the task accurately. This result is contrast to Diffusion Policy, which is extremely sensitive to the distractor. We observe that the Diffusion Policy tends to target the central points of objects, indicating a deficiency in adapting to new environments. In contrast, our model demonstrates greater robustness to changes in the scene, suggesting superior adaptability.

\section{Conclusion}
\label{sec:conclusion}
In this study, we explore the transformer architecture within the context of Diffusion Policy. We pinpoint the large gradient of condition fusion as the fundamental challenge in transformer architecture. Our proposed architecture facilitates training with increased model sizes up to one billion parameters. We present a preliminary study indicating that incorporating a greater number of parameters into the diffusion transformer policy model enables the emergence of properties not observed in smaller-scale models. Our method presents the first attempt to scale up model size for diffusion-based imitation learning, which we believe will be an important direction for future research.

\section{Acknowledgements}
We sincerely thank Yanjie Ze for contributions in discussions and paper review. Our gratitude also thank Yong Wang for his assistance with the hardware setup.

\bibliographystyle{IEEEtran}
\bibliography{main}

\begin{thebibliography}{10}
\providecommand{\url}[1]{#1}
\csname url@samestyle\endcsname
\providecommand{\newblock}{\relax}
\providecommand{\bibinfo}[2]{#2}
\providecommand{\BIBentrySTDinterwordspacing}{\spaceskip=0pt\relax}
\providecommand{\BIBentryALTinterwordstretchfactor}{4}
\providecommand{\BIBentryALTinterwordspacing}{\spaceskip=\fontdimen2\font plus
\BIBentryALTinterwordstretchfactor\fontdimen3\font minus \fontdimen4\font\relax}
\providecommand{\BIBforeignlanguage}[2]{{%
\expandafter\ifx\csname l@#1\endcsname\relax
\typeout{** WARNING: IEEEtran.bst: No hyphenation pattern has been}%
\typeout{** loaded for the language `#1'. Using the pattern for}%
\typeout{** the default language instead.}%
\else
\language=\csname l@#1\endcsname
\fi
#2}}
\providecommand{\BIBdecl}{\relax}
\BIBdecl

\bibitem{DDPM}
J.~Ho, A.~Jain, and P.~Abbeel, ``Denoising diffusion probabilistic models,'' \emph{Advances in neural information processing systems}, vol.~33, pp. 6840--6851, 2020.

\bibitem{DDIM}
J.~Song, C.~Meng, and S.~Ermon, ``Denoising diffusion implicit models,'' \emph{arXiv preprint arXiv:2010.02502}, 2020.

\bibitem{classifierddpm}
J.~Ho and T.~Salimans, ``Classifier-free diffusion guidance,'' \emph{arXiv preprint arXiv:2207.12598}, 2022.

\bibitem{Imagen}
C.~Saharia, W.~Chan, S.~Saxena, L.~Li, J.~Whang, E.~L. Denton, K.~Ghasemipour, R.~Gontijo~Lopes, B.~Karagol~Ayan, T.~Salimans \emph{et~al.}, ``Photorealistic text-to-image diffusion models with deep language understanding,'' \emph{Advances in neural information processing systems}, vol.~35, pp. 36\,479--36\,494, 2022.

\bibitem{richter2023speech}
J.~Richter, S.~Welker, J.-M. Lemercier, B.~Lay, and T.~Gerkmann, ``Speech enhancement and dereverberation with diffusion-based generative models,'' \emph{IEEE/ACM Transactions on Audio, Speech, and Language Processing}, 2023.

\bibitem{sora}
\BIBentryALTinterwordspacing
T.~Brooks, B.~Peebles, C.~Holmes, W.~DePue, Y.~Guo, L.~Jing, D.~Schnurr, J.~Taylor, T.~Luhman, E.~Luhman, C.~Ng, R.~Wang, and A.~Ramesh, ``Video generation models as world simulators,'' 2024. [Online]. Available: \url{https://openai.com/research/video-generation-models-as-world-simulators}
\BIBentrySTDinterwordspacing

\bibitem{imagenvideo}
J.~Ho, W.~Chan, C.~Saharia, J.~Whang, R.~Gao, A.~Gritsenko, D.~P. Kingma, B.~Poole, M.~Norouzi, D.~J. Fleet \emph{et~al.}, ``Imagen video: High definition video generation with diffusion models,'' \emph{arXiv preprint arXiv:2210.02303}, 2022.

\bibitem{lee2024dreamflow}
\BIBentryALTinterwordspacing
K.~Lee, K.~Sohn, and J.~Shin, ``Dreamflow: High-quality text-to-3d generation by approximating probability flow,'' in \emph{The Twelfth International Conference on Learning Representations}, 2024. [Online]. Available: \url{https://openreview.net/forum?id=GURqUuTebY}
\BIBentrySTDinterwordspacing

\bibitem{chi2023diffusion_policy}
C.~Chi, S.~Feng, Y.~Du, Z.~Xu, E.~Cousineau, B.~Burchfiel, and S.~Song, ``Diffusion policy: Visuomotor policy learning via action diffusion,'' \emph{RSS}, 2023.

\bibitem{prasad2024consistency}
A.~Prasad, K.~Lin, J.~Wu, L.~Zhou, and J.~Bohg, ``Consistency policy: Accelerated visuomotor policies via consistency distillation,'' \emph{arXiv preprint arXiv:2405.07503}, 2024.

\bibitem{touvron2023llama}
H.~Touvron, L.~Martin, K.~Stone, P.~Albert, A.~Almahairi, Y.~Babaei, N.~Bashlykov, S.~Batra, P.~Bhargava, S.~Bhosale \emph{et~al.}, ``Llama 2: Open foundation and fine-tuned chat models,'' \emph{arXiv preprint arXiv:2307.09288}, 2023.

\bibitem{dehghani2023scaling}
M.~Dehghani, J.~Djolonga, B.~Mustafa, P.~Padlewski, J.~Heek, J.~Gilmer, A.~P. Steiner, M.~Caron, R.~Geirhos, I.~Alabdulmohsin \emph{et~al.}, ``Scaling vision transformers to 22 billion parameters,'' in \emph{International Conference on Machine Learning}.\hskip 1em plus 0.5em minus 0.4em\relax PMLR, 2023, pp. 7480--7512.

\bibitem{ha2023scaling}
H.~Ha, P.~Florence, and S.~Song, ``Scaling up and distilling down: Language-guided robot skill acquisition,'' in \emph{Conference on Robot Learning}.\hskip 1em plus 0.5em minus 0.4em\relax PMLR, 2023, pp. 3766--3777.

\bibitem{zhai2022scaling}
X.~Zhai, A.~Kolesnikov, N.~Houlsby, and L.~Beyer, ``Scaling vision transformers,'' in \emph{Proceedings of the IEEE/CVF conference on computer vision and pattern recognition}, 2022, pp. 12\,104--12\,113.

\bibitem{tranformer}
A.~Vaswani, N.~Shazeer, N.~Parmar, J.~Uszkoreit, L.~Jones, A.~N. Gomez, {\L}.~Kaiser, and I.~Polosukhin, ``Attention is all you need,'' \emph{Advances in neural information processing systems}, vol.~30, 2017.

\bibitem{vit}
\BIBentryALTinterwordspacing
A.~Dosovitskiy, L.~Beyer, A.~Kolesnikov, D.~Weissenborn, X.~Zhai, T.~Unterthiner, M.~Dehghani, M.~Minderer, G.~Heigold, S.~Gelly, J.~Uszkoreit, and N.~Houlsby, ``An image is worth 16x16 words: Transformers for image recognition at scale,'' in \emph{International Conference on Learning Representations}, 2021. [Online]. Available: \url{https://openreview.net/forum?id=YicbFdNTTy}
\BIBentrySTDinterwordspacing

\bibitem{kaplan2020scaling}
J.~Kaplan, S.~McCandlish, T.~Henighan, T.~B. Brown, B.~Chess, R.~Child, S.~Gray, A.~Radford, J.~Wu, and D.~Amodei, ``Scaling laws for neural language models,'' \emph{arXiv preprint arXiv:2001.08361}, 2020.

\bibitem{yu2020metaworld}
T.~Yu, D.~Quillen, Z.~He, R.~Julian, K.~Hausman, C.~Finn, and S.~Levine, ``Meta-world: A benchmark and evaluation for multi-task and meta reinforcement learning,'' in \emph{Conference on robot learning}.\hskip 1em plus 0.5em minus 0.4em\relax PMLR, 2020, pp. 1094--1100.

\bibitem{karras2019style}
T.~Karras, S.~Laine, and T.~Aila, ``A style-based generator architecture for generative adversarial networks,'' in \emph{Proceedings of the IEEE/CVF conference on computer vision and pattern recognition}, 2019, pp. 4401--4410.

\bibitem{stablediffusion}
R.~Rombach, A.~Blattmann, D.~Lorenz, P.~Esser, and B.~Ommer, ``High-resolution image synthesis with latent diffusion models,'' in \emph{Proceedings of the IEEE/CVF conference on computer vision and pattern recognition}, 2022, pp. 10\,684--10\,695.

\bibitem{yang2023policy}
L.~Yang, Z.~Huang, F.~Lei, Y.~Zhong, Y.~Yang, C.~Fang, S.~Wen, B.~Zhou, and Z.~Lin, ``Policy representation via diffusion probability model for reinforcement learning,'' \emph{arXiv preprint arXiv:2305.13122}, 2023.

\bibitem{mazoure2023value}
B.~Mazoure, W.~Talbott, M.~A. Bautista, D.~Hjelm, A.~Toshev, and J.~Susskind, ``Value function estimation using conditional diffusion models for control,'' \emph{arXiv preprint arXiv:2306.07290}, 2023.

\bibitem{brehmer2024edgi}
J.~Brehmer, J.~Bose, P.~De~Haan, and T.~S. Cohen, ``Edgi: Equivariant diffusion for planning with embodied agents,'' \emph{Advances in Neural Information Processing Systems}, vol.~36, 2024.

\bibitem{venkatraman2023reasoning}
S.~Venkatraman, S.~Khaitan, R.~T. Akella, J.~Dolan, J.~Schneider, and G.~Berseth, ``Reasoning with latent diffusion in offline reinforcement learning,'' \emph{arXiv preprint arXiv:2309.06599}, 2023.

\bibitem{lee2024refining}
K.~Lee, S.~Kim, and J.~Choi, ``Refining diffusion planner for reliable behavior synthesis by automatic detection of infeasible plans,'' \emph{Advances in Neural Information Processing Systems}, vol.~36, 2024.

\bibitem{zhou2024adaptive}
S.~Zhou, Y.~Du, S.~Zhang, M.~Xu, Y.~Shen, W.~Xiao, D.-Y. Yeung, and C.~Gan, ``Adaptive online replanning with diffusion models,'' \emph{Advances in Neural Information Processing Systems}, vol.~36, 2024.

\bibitem{chen2022offline}
H.~Chen, C.~Lu, C.~Ying, H.~Su, and J.~Zhu, ``Offline reinforcement learning via high-fidelity generative behavior modeling,'' \emph{arXiv preprint arXiv:2209.14548}, 2022.

\bibitem{ze2023visual}
Y.~Ze, N.~Hansen, Y.~Chen, M.~Jain, and X.~Wang, ``Visual reinforcement learning with self-supervised 3d representations,'' \emph{IEEE Robotics and Automation Letters}, vol.~8, no.~5, pp. 2890--2897, 2023.

\bibitem{khazatsky2024droid}
A.~Khazatsky, K.~Pertsch, S.~Nair, A.~Balakrishna, S.~Dasari, S.~Karamcheti, S.~Nasiriany, M.~K. Srirama, L.~Y. Chen, K.~Ellis \emph{et~al.}, ``Droid: A large-scale in-the-wild robot manipulation dataset,'' \emph{arXiv preprint arXiv:2403.12945}, 2024.

\bibitem{vosylius2024render}
V.~Vosylius, Y.~Seo, J.~Uru{\c{c}}, and S.~James, ``Render and diffuse: Aligning image and action spaces for diffusion-based behaviour cloning,'' \emph{arXiv preprint arXiv:2405.18196}, 2024.

\bibitem{reuss2024multimodal}
M.~Reuss, {\"O}.~E. Ya{\u{g}}murlu, F.~Wenzel, and R.~Lioutikov, ``Multimodal diffusion transformer: Learning versatile behavior from multimodal goals,'' in \emph{First Workshop on Vision-Language Models for Navigation and Manipulation at ICRA 2024}.

\bibitem{ze20243d}
Y.~Ze, G.~Zhang, K.~Zhang, C.~Hu, M.~Wang, and H.~Xu, ``3d diffusion policy: Generalizable visuomotor policy learning via simple 3d representations,'' in \emph{ICRA 2024 Workshop on 3D Visual Representations for Robot Manipulation}, 2024.

\bibitem{team2024octo}
O.~M. Team, D.~Ghosh, H.~Walke, K.~Pertsch, K.~Black, O.~Mees, S.~Dasari, J.~Hejna, T.~Kreiman, C.~Xu \emph{et~al.}, ``Octo: An open-source generalist robot policy,'' \emph{arXiv preprint arXiv:2405.12213}, 2024.

\bibitem{xian2023unifying}
Z.~Xian, N.~Gkanatsios, T.~Gervet, and K.~Fragkiadaki, ``Unifying diffusion models with action detection transformers for multi-task robotic manipulation,'' in \emph{7th Annual Conference on Robot Learning}, 2023.

\bibitem{ze2023gnfactor}
Y.~Ze, G.~Yan, Y.-H. Wu, A.~Macaluso, Y.~Ge, J.~Ye, N.~Hansen, L.~E. Li, and X.~Wang, ``Gnfactor: Multi-task real robot learning with generalizable neural feature fields,'' in \emph{Conference on Robot Learning}.\hskip 1em plus 0.5em minus 0.4em\relax PMLR, 2023, pp. 284--301.

\bibitem{reuss2023goal}
M.~Reuss, M.~Li, X.~Jia, and R.~Lioutikov, ``Goal-conditioned imitation learning using score-based diffusion policies,'' \emph{arXiv preprint arXiv:2304.02532}, 2023.

\bibitem{pearce2023imitating}
T.~Pearce, T.~Rashid, A.~Kanervisto, D.~Bignell, M.~Sun, R.~Georgescu, S.~V. Macua, S.~Z. Tan, I.~Momennejad, K.~Hofmann \emph{et~al.}, ``Imitating human behaviour with diffusion models,'' \emph{arXiv preprint arXiv:2301.10677}, 2023.

\bibitem{ze2023visual3d}
Y.~Ze, N.~Hansen, Y.~Chen, M.~Jain, and X.~Wang, ``Visual reinforcement learning with self-supervised 3d representations,'' \emph{IEEE Robotics and Automation Letters}, vol.~8, no.~5, pp. 2890--2897, 2023.

\bibitem{zhu2024retrieval}
Y.~Zhu, Z.~Ou, X.~Mou, and J.~Tang, ``Retrieval-augmented embodied agents,'' 2024.

\bibitem{yang2024movie}
S.~Yang, Y.~Ze, and H.~Xu, ``Movie: Visual model-based policy adaptation for view generalization,'' \emph{Advances in Neural Information Processing Systems}, vol.~36, 2024.

\bibitem{psenka2023learning}
M.~Psenka, A.~Escontrela, P.~Abbeel, and Y.~Ma, ``Learning a diffusion model policy from rewards via q-score matching,'' \emph{arXiv preprint arXiv:2312.11752}, 2023.

\bibitem{huang2023diffusion}
T.~Huang, G.~Jiang, Y.~Ze, and H.~Xu, ``Diffusion reward: Learning rewards via conditional video diffusion,'' \emph{arXiv preprint arXiv:2312.14134}, 2023.

\bibitem{ze2024h}
Y.~Ze, Y.~Liu, R.~Shi, J.~Qin, Z.~Yuan, J.~Wang, and H.~Xu, ``H-index: Visual reinforcement learning with hand-informed representations for dexterous manipulation,'' \emph{Advances in Neural Information Processing Systems}, vol.~36, 2024.

\bibitem{urain2023se}
J.~Urain, N.~Funk, J.~Peters, and G.~Chalvatzaki, ``Se (3)-diffusionfields: Learning smooth cost functions for joint grasp and motion optimization through diffusion,'' in \emph{2023 IEEE International Conference on Robotics and Automation (ICRA)}.\hskip 1em plus 0.5em minus 0.4em\relax IEEE, 2023, pp. 5923--5930.

\bibitem{simeonov2023shelving}
A.~Simeonov, A.~Goyal, L.~Manuelli, L.~Yen-Chen, A.~Sarmiento, A.~Rodriguez, P.~Agrawal, and D.~Fox, ``Shelving, stacking, hanging: Relational pose diffusion for multi-modal rearrangement,'' \emph{arXiv preprint arXiv:2307.04751}, 2023.

\bibitem{wu2024learning}
T.~Wu, M.~Wu, J.~Zhang, Y.~Gan, and H.~Dong, ``Learning score-based grasping primitive for human-assisting dexterous grasping,'' \emph{Advances in Neural Information Processing Systems}, vol.~36, 2024.

\bibitem{wen2024tinyvla}
J.~Wen, Y.~Zhu, J.~Li, M.~Zhu, K.~Wu, Z.~Xu, R.~Cheng, C.~Shen, Y.~Peng, F.~Feng \emph{et~al.}, ``Tinyvla: Towards fast, data-efficient vision-language-action models for robotic manipulation,'' \emph{arXiv preprint arXiv:2409.12514}, 2024.

\bibitem{zhong2023language}
Z.~Zhong, D.~Rempe, Y.~Chen, B.~Ivanovic, Y.~Cao, D.~Xu, M.~Pavone, and B.~Ray, ``Language-guided traffic simulation via scene-level diffusion,'' in \emph{Conference on Robot Learning}.\hskip 1em plus 0.5em minus 0.4em\relax PMLR, 2023, pp. 144--177.

\bibitem{liu2023mm}
X.~Liu, Y.~Zhu, J.~Gu, Y.~Lan, C.~Yang, and Y.~Qiao, ``Mm-safetybench: A benchmark for safety evaluation of multimodal large language models,'' \emph{arXiv preprint arXiv:2311.17600}, 2023.

\bibitem{zhu2024llava}
Y.~Zhu, M.~Zhu, N.~Liu, Z.~Ou, X.~Mou, and J.~Tang, ``Llava-phi: Efficient multi-modal assistant with small language model,'' \emph{arXiv preprint arXiv:2401.02330}, 2024.

\bibitem{liu2022structdiffusion}
W.~Liu, T.~Hermans, S.~Chernova, and C.~Paxton, ``Structdiffusion: Object-centric diffusion for semantic rearrangement of novel objects,'' in \emph{Workshop on Language and Robotics at CoRL 2022}, 2022.

\bibitem{saha2023edmp}
K.~Saha, V.~Mandadi, J.~Reddy, A.~Srikanth, A.~Agarwal, B.~Sen, A.~Singh, and M.~Krishna, ``Edmp: Ensemble-of-costs-guided diffusion for motion planning,'' \emph{arXiv preprint arXiv:2309.11414}, 2023.

\bibitem{chang2023denoising}
J.~Chang, H.~Ryu, J.~Kim, S.~Yoo, J.~Seo, N.~Prakash, J.~Choi, and R.~Horowitz, ``Denoising heat-inspired diffusion with insulators for collision free motion planning,'' \emph{arXiv preprint arXiv:2310.12609}, 2023.

\bibitem{wen2024object}
J.~Wen, Y.~Zhu, M.~Zhu, J.~Li, Z.~Xu \emph{et~al.}, ``Object-centric instruction augmentation for robotic manipulation,'' 2024.

\bibitem{zhu2024language}
M.~Zhu, Y.~Zhu, J.~Li, J.~Wen, Z.~Xu \emph{et~al.}, ``Language-conditioned robotic manipulation with fast and slow thinking,'' 2024.

\bibitem{welling2011bayesian}
M.~Welling and Y.~W. Teh, ``Bayesian learning via stochastic gradient langevin dynamics,'' in \emph{Proceedings of the 28th international conference on machine learning (ICML-11)}.\hskip 1em plus 0.5em minus 0.4em\relax Citeseer, 2011, pp. 681--688.

\bibitem{zhao2022penalizing}
Y.~Zhao, H.~Zhang, and X.~Hu, ``Penalizing gradient norm for efficiently improving generalization in deep learning,'' in \emph{International Conference on Machine Learning}.\hskip 1em plus 0.5em minus 0.4em\relax PMLR, 2022, pp. 26\,982--26\,992.

\bibitem{zhou2020towards}
Y.~Zhou, B.~Karimi, J.~Yu, Z.~Xu, and P.~Li, ``Towards better generalization of adaptive gradient methods,'' \emph{Advances in Neural Information Processing Systems}, vol.~33, pp. 810--821, 2020.

\bibitem{pmlr-v139-akbari21a}
\BIBentryALTinterwordspacing
A.~Akbari, M.~Awais, M.~Bashar, and J.~Kittler, ``How does loss function affect generalization performance of deep learning? application to human age estimation,'' in \emph{Proceedings of the 38th International Conference on Machine Learning}, ser. Proceedings of Machine Learning Research, M.~Meila and T.~Zhang, Eds., vol. 139.\hskip 1em plus 0.5em minus 0.4em\relax PMLR, 18--24 Jul 2021, pp. 141--151. [Online]. Available: \url{https://proceedings.mlr.press/v139/akbari21a.html}
\BIBentrySTDinterwordspacing

\bibitem{peebles2023scalable}
W.~Peebles and S.~Xie, ``Scalable diffusion models with transformers,'' in \emph{Proceedings of the IEEE/CVF International Conference on Computer Vision}, 2023, pp. 4195--4205.

\bibitem{brock2018large}
A.~Brock, J.~Donahue, and K.~Simonyan, ``Large scale gan training for high fidelity natural image synthesis,'' \emph{arXiv preprint arXiv:1809.11096}, 2018.

\bibitem{act}
T.~Z. Zhao, V.~Kumar, S.~Levine, and C.~Finn, ``Learning fine-grained bimanual manipulation with low-cost hardware,'' \emph{arXiv preprint arXiv:2304.13705}, 2023.

\bibitem{seo2023masked}
Y.~Seo, D.~Hafner, H.~Liu, F.~Liu, S.~James, K.~Lee, and P.~Abbeel, ``Masked world models for visual control,'' in \emph{Conference on Robot Learning}.\hskip 1em plus 0.5em minus 0.4em\relax PMLR, 2023, pp. 1332--1344.

\end{thebibliography}

\end{document}